\documentclass[12pt]{article}
\usepackage{fullpage}
\usepackage{graphicx}
\usepackage{amssymb}
\usepackage{amsmath}
\usepackage{theorem}
\usepackage{natbib}

\title{Forest Garrote}
\author{Nicolai Meinshausen\footnote{Department of Statistics, University of Oxford, 1 South Parks Road, OX1 3TG, UK}}

\begin{document}
\maketitle
\begin{abstract} 
Variable selection for high-dimensional linear models has received a lot of attention lately, mostly in the context of $\ell_1$-regularization. Part of the attraction is the variable selection effect: parsimonious models are obtained, which are very suitable for interpretation. In terms of predictive power, however, these regularized linear models are  often slightly inferior to machine learning procedures like tree ensembles. Tree ensembles, on the other hand, lack usually a formal way of variable selection and are difficult to visualize. A Garrote-style convex penalty for trees ensembles, in particular Random Forests, is proposed. The penalty selects functional groups of nodes in the trees. These could be as simple as monotone functions of individual predictor variables. This yields a parsimonious function fit, which lends itself easily to visualization and interpretation. The predictive power is maintained at least at the same level as the original tree ensemble. A key feature of the method is that, once a tree ensemble is fitted, no further tuning parameter needs to be selected. The empirical performance is demonstrated on a wide array of datasets.
\end{abstract}

\section{Introduction}
Given data $(X_i,Y_i)$, for $i=1,\ldots,n$, with a $p$-dimensional real-valued predictor variable $X$, where $X=(X^{(1)},\ldots,X^{(p)})\in\mathcal{X}$, and a real-valued response $Y$, a typical goal of regression analysis is to find an estimator $\hat{Y}(x)$, such that the expected loss $E\big(L(\hat{Y}(X),X)\big)$ is minimal, under a given loss function $L:\mathcal{X}\times \mathbb{R}\mapsto \mathbb{R}^+$. For the following, the standard squared error loss is used. If the predictor can be of the `black-box' type, tree ensembles have proven to be very powerful. Random Forests \citep{breiman01random} is a prime example, as are boosted regression trees \citep{yu2003blr}. There are many interesting tools available for interpretation of these tree ensembles, see for example \citet{strobl2007brf} and the references therein. 

While tree ensembles often have very good predictive performance, an advantage of a linear model is better interpretability. Measuring variable importance and performing variable selection are more easier to formulate and understand in the context of linear models. For high-dimensional data with $p\gg n$, regularization is clearly imperative and the Lasso \citep{tibshirani96regression,chen01atomic} has proven to be very popular in recent years, since it combines a convex optimization problem with variable selection. 
A precursor to the Lasso was the nonnegative Garrote \citep{breiman95better}. A disadvantage of the nonnegative Garrote is the reliance on an initial estimator, which could be the least squares estimator or a regularized variation. On the positive side, important variables incur less penalty and bias under the regularization than they do with the Lasso. For a deeper discussion of the properties of the nonnegative Garrote see \citet{yuan05on}. 

Here, it is proposed to use Random Forest as an initial estimator for the nonnegative Garrote. The idea is related to the Rule Ensemble approach of 
\citet{friedman2005plv}, who used the Lasso instead of the nonnegative Garrote. A crucial distinction is that rules  fulfilling the same functional role are grouped  in our approach. This is similar in spirit to the group Lasso \citep{meier06group,yuan05msa,zhao06cap}. This produces a very accurate predictor that uses just a few functional groups of rules, discarding many variables in the process as irrelevant. 

A unique feature of the proposed method is that is seems to work very well in the absence of a tuning parameter. It just requires the choice of an initial tree ensemble. This makes the procedure very simple to implement and computationally efficient. The idea and the algorithm is developed in Section~\ref{section:2}, while a detailed numerical study on 15 datasets makes up Section~\ref{section:3}.

\section{Methods}
\label{section:2}
\subsection{Trees and Equivalent Rules}
A tree $T$ is seen here as a piecewise-constant function $\mathbb{R}^p\mapsto\mathbb{R}$ derived from the tree structure in the sense of \cite{CART}.
\citet{friedman2005plv} proposed `rules' as a name for simple rectangular-shaped indicator functions.  
Every node $j$ in a tree is associated with a $B_j$ in $\mathbb{R}^p$-dimensional space, defined as the set of all values $x\in\mathbb{R}^p$ that pass through node $j$ if passed down the tree. All values $x\in\mathbb{R}^p$ that do not pass throuh node $j$ are outside of $B_j$. The way rules are used here, they correspond to indicator functions $R=R_{j}$,
\[ R_{j}(x)= \left\{ \begin{array}{cc} 1 & X\in B_j \\ 0& X\notin B_j     \end{array}  \right. ,\]
i.e.\ $R_j(x)=1\{ X\in B_j\}$ is the indicator function for box $B_j$. For a more detailed discussion see \citet{friedman2005plv}. 

To give an example of a rule, take the well-know dataset on abalone \citep{abalone} as an example. The goal is to predict age of abalone from physical measurements. For each of the 4177 abalone in the dataset, eight predictor variables (sex, length, diameter, height, while weight, shucked weight, viscera weight and shell weight) are available. 
An example of a rule is 
\begin{equation}\label{ruleex} R_1(x) = 1\left\{ \mbox{diameter} \ge 0.537 \mbox{  and  shell weight}  \ge 0.135  \right\},\end{equation}
and the presence of such a rule in a final predictor is  easy to interpret, comparable to interpreting coefficients in a linear model.
For the following, it is assumed that rules contain at most a single inequality for each variable. In other words, the boxes defined by rules in $p$-dimensional spaces are defined by at most a single hyperplane in each variable. If a rule violates this assumption, it can easily be decomposed into two or several rules satisfying the assumption.

Every regression tree can be written as a linear superposition of rules. Suppose a tree $T$ has $J$ nodes in total. The regression function $\hat{T}$ of this tree (ensemble) can then be written as 
\begin{equation}\label{treehat} \hat{T}(x) = \sum_{j=1}^J \hat{\beta}_j^{tree} R_{j} (x) \end{equation} for some $\hat{\beta}^{tree}$. The decomposition is not unique in general.  We could, for example, assign non-zero regression coefficients $\hat{\beta}_j$ only to leaf nodes. Here, we build the regression function incrementally instead, assigning non-zero regression coefficients to \emph{all} nodes.  The value $\hat{\beta}_j^{tree}$ are defined as
\begin{equation}\label{eq:hatbetatree} \hat{\beta}_j^{tree} = \left\{ \begin{array}{ll}  E_n (Y|Y\in B_j) - E_n (Y|Y\in B_{pa(j)}) & \mbox{if } j \mbox{ is not root node} \\ E_n(Y) & \mbox{if } j \mbox{ is root node} \end{array} \right. ,\end{equation} where $E_n$ is the empirical mean across the $n$ observations and $pa(j)$ is the parent node of $j$ in the tree.  
Rule (\ref{ruleex}),  in the abalone example above, receives a regression weight $\hat{\beta}_1^{tree}=0.0237$. The contribution of rule (\ref{ruleex}) to the Random Forest fit is thus to increase the fitted value if and only if the diameter is larger than 0.537 and shell weight is larger than 0.135. 
The Random Forest fit is the sum of the contribution from all these rules, where each rule corresponds to one node in the tree ensemble.

To see that (\ref{treehat}) and (\ref{eq:hatbetatree}) really correspond to the original tree (ensemble) solution, consider just a single tree for the moment. 
Denote the predicted value for predictor variable $X$ by $\hat{T}(x)$. Let $B_{\mathit{leaf}(x)}$ be the rectangular area in $p$-dimensional space that corresponds to the leaf node $\mathit{leaf}(x)$ of the tree in which predictor variable $X$ falls. The predicted value follows then by adding up all relevant nodes and obtaining with (\ref{treehat}) and (\ref{eq:hatbetatree}),
\[ \hat{T}(x) = E_n(Y| Y\in B_{\mathit{leaf}(x)}),\]
i.e.\ the predicted values is just the empirical mean of $Y$ across all observations $i=1,\ldots,n$ which fall into the same leaf node as $X$, which is equivalent to the original prediction of the tree. The equivalence for tree ensembles follows by averaging across all individual trees as in (\ref{treehat}).

\subsection{Rule Ensembles}
The idea of \cite{friedman2005plv} is to modify the coefficients $\hat{\beta}^{tree}$ in (\ref{treehat}), increasing sparsity of the fit by setting many regression coefficient to 0 and eliminating the corresponding rules from the fit, while hopefully not degrading the predictive performance of the tree ensemble in the process. The rule ensemble predictors are thus of the form
\begin{equation}\label{hatY} \sum_{j=1}^J  \hat{\beta}_j R_j(x).\end{equation}
Sparsity is enforced by penalizing the $\ell_1$-norm of $\hat{\beta}$ in LASSO-style \citep{tibshirani96regression}, 
\begin{equation}\label{ruleens} \hat{\beta}^{re;\lambda} = \mbox{argmin}_\beta  \; \sum_{i=1}^n (Y_i -  \sum_{j=1}^J  \beta_j R_j(X_i))^2 \quad \mbox{such that  } \sum_{j=1}^J  |\beta_j| \le \lambda.    \end{equation}
This enforces sparsity in terms of rules, i.e.\ the final predictor will have typically only very few rules, at least compared to the original tree ensemble.  
The penalty parameter $\lambda$ is typically chosen by cross-validation. 
\citet{friedman2005plv} recommend to add the linear main effects of all variables into (\ref{ruleens}), which were omitted here for notational simplicity and to keep invariance with respect to monotone transformations of predictor variables.
It is shown in \citet{friedman2005plv} that the rule ensemble estimator maintains in general the predictive ability of Random Forests, while lending itself more easily to interpretation.

\subsection{Functional grouping of rules}

\begin{figure}
\includegraphics[width=0.25\textwidth]{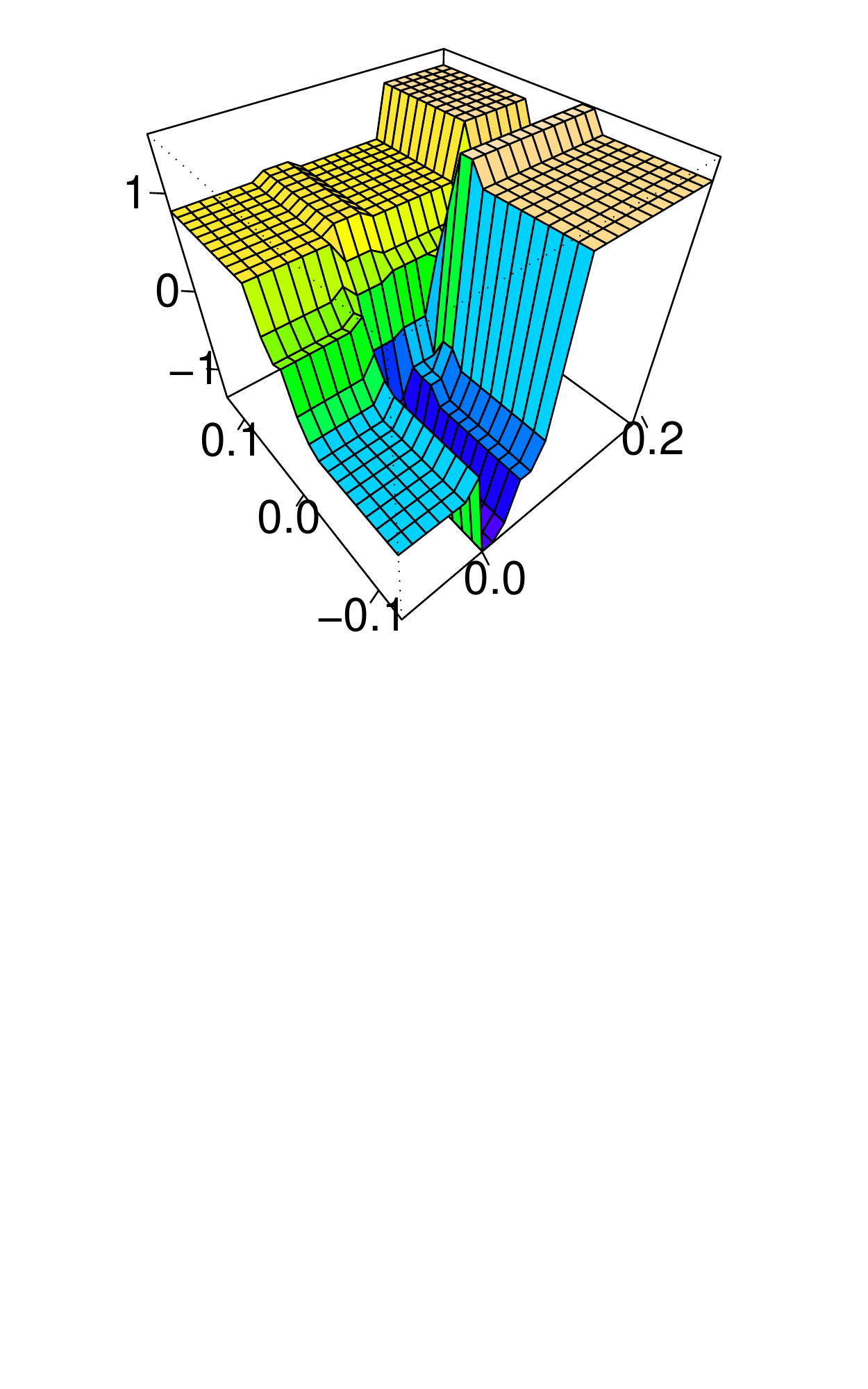}
\includegraphics[width=0.24\textwidth]{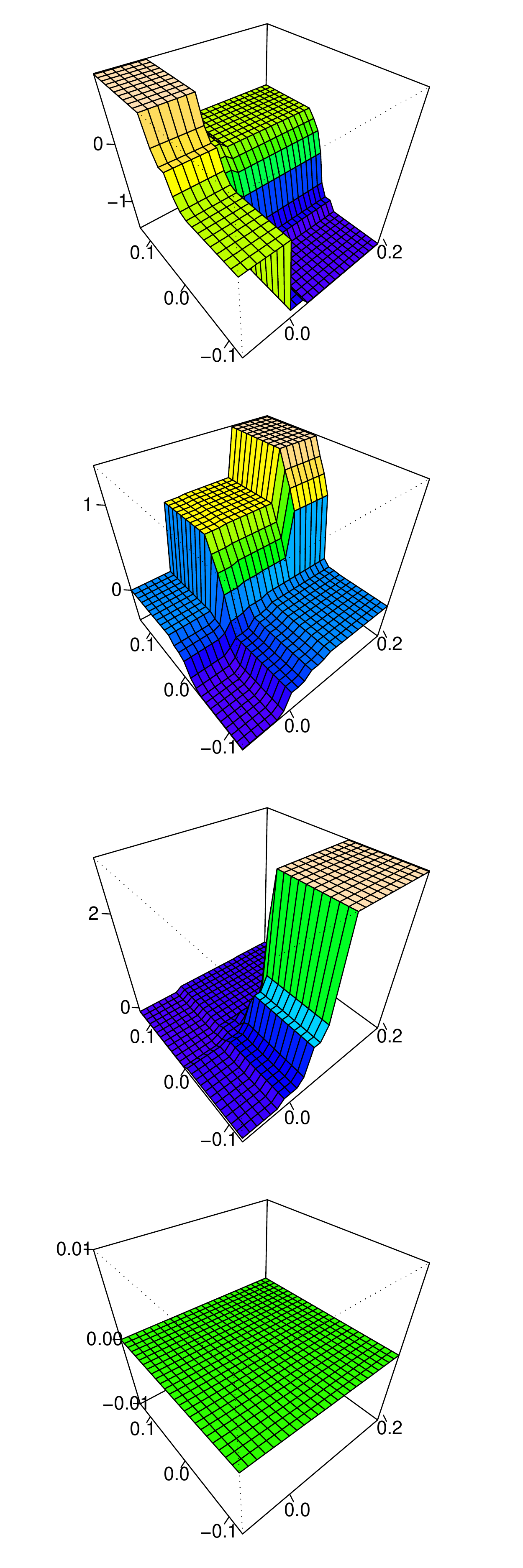}
\includegraphics[width=0.24\textwidth]{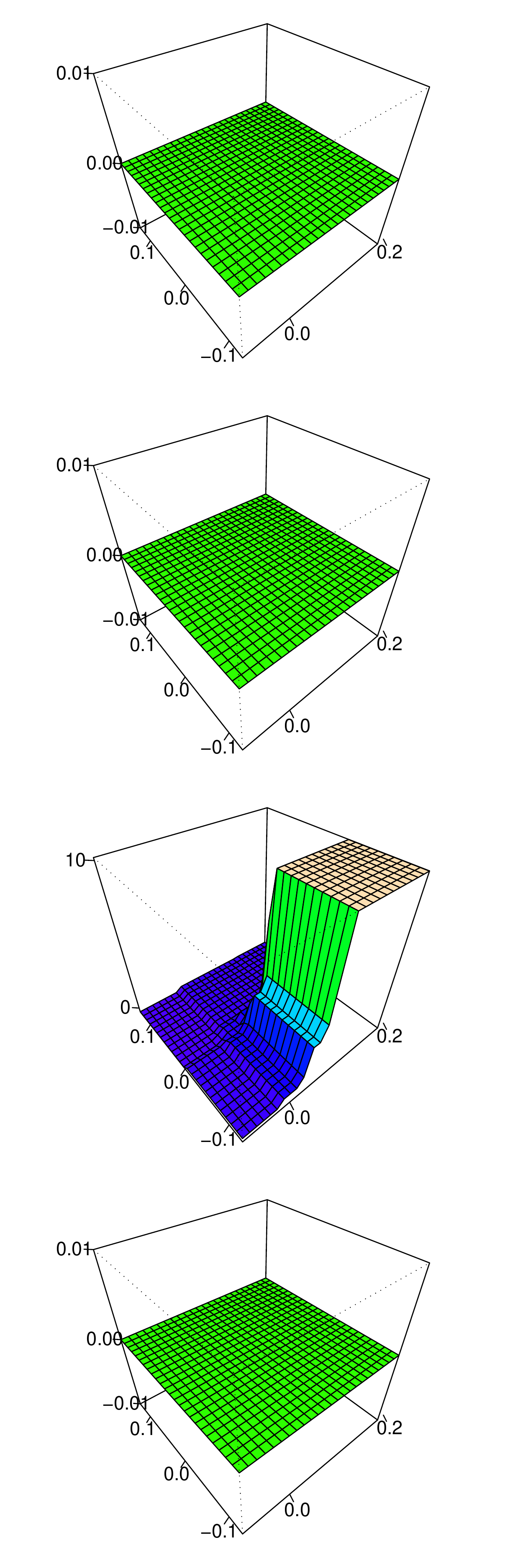}
\includegraphics[width=0.25\textwidth]{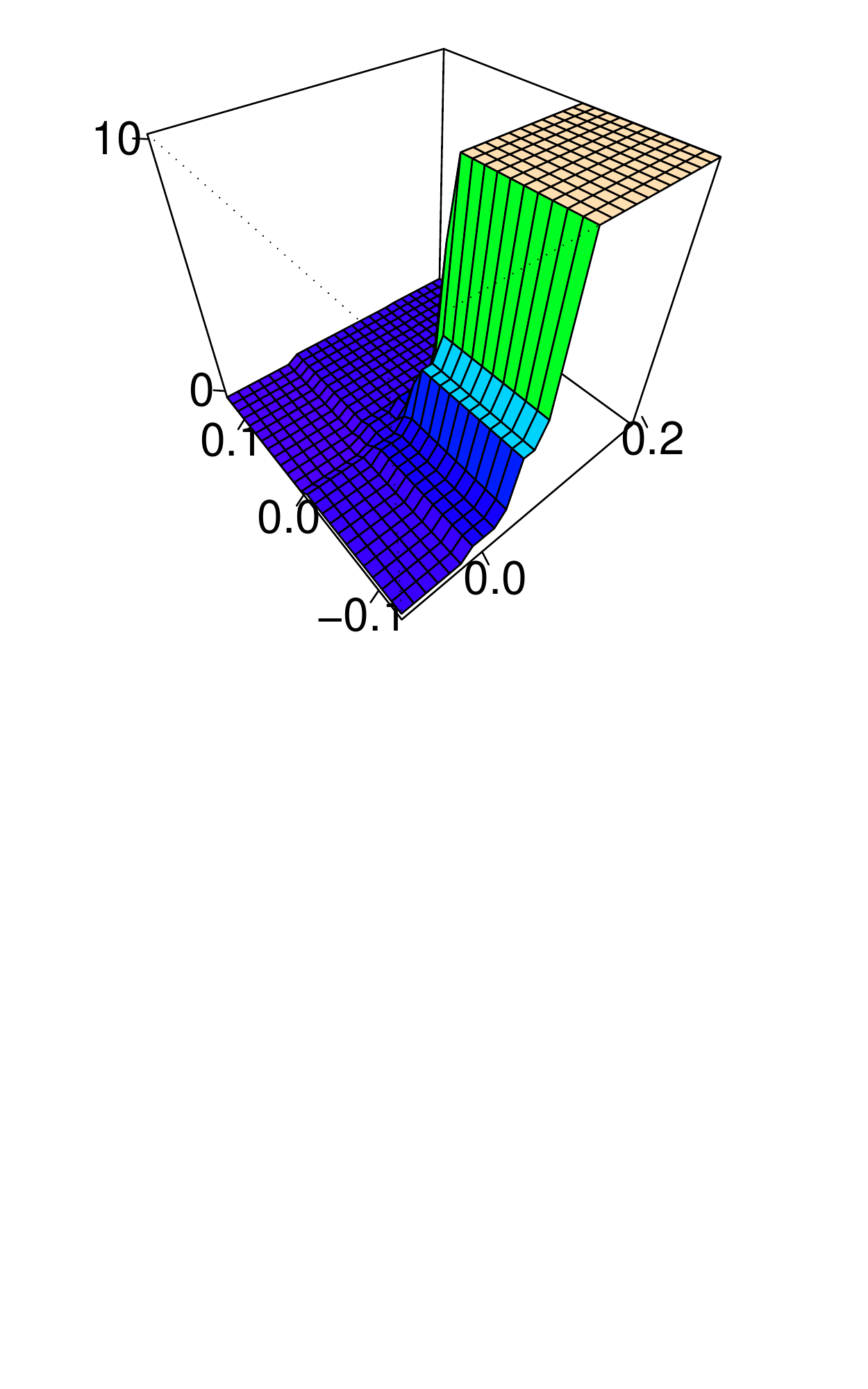}

\caption{ \textit{\label{fig:38} \textit{ FIRST COLUMN:  combined variable interaction rules between predictor variable \emph{tch} (x-axis to the right) and \emph{ltg} (y-axis to the left) in the Random Forest fit for the Diabetes data.  SECOND COLUMN: the interaction rules can be decomposed into the effects $\hat{T}_\sigma$ (plotted on z-axis) of four interaction patterns $\sigma=(+,-)$ on top, $(+,+)$ on second from top, $(-,+)$ on second from bottom and $(-,-)$ on the bottom. THIRD COLUMN: applying the Garrote correction, three of the four interaction patterns are set to 0. FOURTH COLUMN: adding the four interaction patterns with Garrote correction up, the total interaction pattern between the two variables \emph{ltg} and \emph{tch} in the Forest Garrote fit.  }}}
\end{figure}

The rule ensemble approach is treating all rules equally by enforcing a $\ell_1$-penalty on all rules extracted from a tree ensemble. It does not take into account, however, that there are typically many very closely related rules in the fit. Take the RF fit to the abalone data as an example. Several hundred rules are extracted from the RF, two of which are 
\begin{equation}\label{ruleex1} R_1(x) = 1\left\{ \mbox{diameter} \ge 0.537 \mbox{  and  shell weight}  \ge 0.135  \right\} ,\end{equation}
with regression coefficient $\hat{\beta}_1=0.023$ and 
\begin{equation}\label{ruleex2} R_2(x) = 1\left\{ \mbox{diameter} \ge 0.537 \mbox{  and  shell weight}  \ge 0.177  \right\} ,\end{equation}
with regression coefficient $\hat{\beta}_2=0.019$.
The effect of these two rules, measured by  $\hat{\beta}_1 R_1$ and $\hat{\beta}_2 R_2$, are clearly very similar.  
In total, there are 32 rules `interaction' rules that involve variables diameter and shell weight in the RF fit to the abalone data. Selecting some members of this group, it seems artificial to exclude others of the same `functional' type.

Sparsity is measured purely on a rule-by-rule basis in the $\ell_1$-penalty term of rule ensembles (\ref{ruleens}). Selecting the two rules mentioned above incurs the same sparsity penalty as if the second rule involved two completely different variables.
An undesirable side-effect of not taking into account the grouping of rules is that many or even all original predictor variables might still be involved in the rules; sparsity is not explicitly enforced in the sense that many irrelevant original predictor variables are completely discarded in the selected rules. 

It seems natural to let rules form functional groups.
The question then turns up which rules form useful and interpretable groups. There is clearly no simple right or wrong answer to this question. Here, a very simple yet hopefully intuitive functional grouping of rules is employed.  For the $j$-th rule, with coefficient $\hat{\beta}_j$, define the interaction pattern $\sigma_{j}=(\sigma_{j,1},\ldots,\sigma_{j,p})$ for variables $k=1,\ldots,p$ by

\begin{equation}\label{sigma} \sigma_{j,k} =\left\{ \begin{array}{cl} +1 & \mbox{ iff   } \sup_{x,x'\in\mathbb{R}^p: x^{(k)}> x'^{(k)}}\; \hat{\beta}_j \; (R_j(x) - R_j(x')) \;>\; 0 \\ -1 & \mbox{ iff   } \inf_{x,x'\in\mathbb{R}^p: x^{(k)}> x'^{(k)}}\;\; \hat{\beta}_j \; (R_j(x) - R_j(x')) \; <\; 0  \\ 0 & \mbox{ otherwise}  \end{array}\right. .\end{equation}
 The meaning of interaction patterns is best understood if looking at examples for varying degrees, where the degree of a interaction pattern $\sigma$ is understood to be the number of non-zero entries in $\sigma$ and corresponds to the number of variables that are involved in a rule.

\paragraph{First degree (main effects).}
The simplest interaction patterns are those involving a single variable only, which correspond in some sense to the main effects of variables. The interaction pattern 
\[ (0,+,0,0,0,0,0,0)\]
for example collects all rules that involve the 2nd predictor variable \textit{length} only and  lead to a monotonically increasing fit (are thus of the form $ 1\{ \mbox{length} \le u \}$ for some real-valued $u$ if the corresponding regression coefficient were positive or $ 1\{ \mbox{length} \ge u\}$ if the regression coefficient were negative). The interaction pattern $(0,0,-,0,0,0,0,0)$ collects conversely all those rules that yield a monotonically decreasing fit in the variable \textit{diameter}, the third variable.

\paragraph{Second degree (interactions effects).}
Second degree interaction patterns are of the form (\ref{ruleex1}) or (\ref{ruleex2}). 
As diameter is the 3rd variable and shell weight the 8th, the interaction pattern of both rules (\ref{ruleex1}) and (\ref{ruleex2}) is 
\[ (0,0,+,0,0,0,0,+),\]
making them members of the same functional group, as for both rules the fitted value is monotonically increasing in both involved variables. In other words, either a large value in both variables increases the fitted value or a very low value in both variables decreases the fitted value.  Second degree interaction patterns thus form four categories for each pair of variables. A case could be made to merge these four categories into just two categories, as the interaction patterns do not conform nicely with the more standard multiplicative form of interactions in linear models. However, there is no reason to believe that nature always adheres to the multiplicative form of interactions typically assumed in linear models. The interaction patterns used here seemed more adapted to the context of rule-based inference.
Factorial variables can be dealt with in the same framework (\ref{sigma}) by converting to dummy variables first. 

\subsection{Garrote correction and selection}
In contrast to the group Lasso approach of \citet{yuan05msa}, the proposed method does not only start with knowledge of natural groups of variables or rules. A very good initial estimator is available, namely the Random Forest fit. This is exploited in the following. 

 Let $\hat{T}_\sigma$ be the part of the fit that collects all contributions from rules with interaction pattern $\sigma$,
\begin{equation}\label{Tsigma} \hat{T}_\sigma (x) \;=  \sum_{j: \sigma(\hat{\beta}_jR_j)=\sigma} \hat{\beta}_j R_j(x)  .  \end{equation}
Let $\mathcal{G}$ be the collection of all possible interaction patterns $\sigma$. The tree ensemble fit (\ref{treehat}) can then be re-written as a sum over all interaction patterns
\begin{equation}\label{Tdecomp} \hat{T}(x) \;\;= \;\; \sum_{\sigma\in\mathcal{G}} \hat{T}_\sigma(x).\end{equation}
A interaction pattern $\sigma$ is called \emph{active} if the corresponding fit in the tree ensemble is non-zero, i.e.\ if and only if $\hat{T}_\sigma$ is not identically  $0$.
The Random Forest fit contains very often a huge number of active interaction pattern, involving interactions up to fourth and higher degrees.
Most of those active patterns contribute just in a negligible way to the overall fit.

The idea proposed here is to use (\ref{Tdecomp}) as a starting point and modify it to enforce sparsity in the final fit, getting rid of as many unnecessary predictor variables and associated interaction patterns as possible.  
The Lasso of \citet{tibshirani96regression} was used in the rule ensemble approach of \citet{friedman2005plv}. Here, however, the  starting point is the functional decomposition (\ref{Tdecomp}), which is already a very good initial (yet not sparse) estimator of the underlying regression function. 

Hence it seems more appropriate to use Breiman's nonnegative Garrote \citep{breiman95better}, penalizing contributions of interaction patterns less if their contribution to the initial estimator is large and vice versa. The beneficial effect of this bias reduction for important variables has been noted in, amongst others, \citet{yuan05on} and \citet{zou05adaptive}.

The Garrote-style Forest estimator $\hat{T}^{gar}$ is defined as
\begin{equation}\label{hatgar} \hat{T}^{gar} \;=\; \sum_{\sigma\in\mathcal{G}} \gamma_\sigma \, \hat{T}_\sigma.  \end{equation}
Each contribution $\hat{T}_\sigma$ of an interaction pattern $\sigma$ is multiplied by a factor $\hat{\gamma}_\sigma$. 
The original tree ensemble fit is obtained by setting all factors equal to 1. 

The multiplicative factor $\gamma$ is chosen by least squares, subject to the constraint that the total $\ell_1$-norm of the multiplying coefficients is less than 1,
\begin{eqnarray}\hat{\gamma} & =& \mbox{argmin}_{\gamma}\; \sum_{i=1}^n (Y_i -  \sum_{\sigma\in\mathcal{G}} \gamma_\sigma  \hat{T}_\sigma(X_i))^2 \nonumber \\ &&  \mbox{     such that } \;\;|\mathcal{G}|^{-1} \sum_{\sigma\in\mathcal{G}} |\gamma_\sigma| \le 1 \;\;\mbox{ and } \min_{\sigma\in\mathcal{G}} \gamma_\sigma \ge 0.\label{gammahat}  \end{eqnarray}
The normalizing factor $|\mathcal{G}|^{-1}$ divides the $\ell_1$-norm of $\gamma$ by the total number of interaction patterns and is certainly not crucial here but simplifies notation. The estimation of $\hat{\gamma}$ is an application of Breiman's nonnegative Garrote \citep{breiman95better}. As for the Garrote, the original predictors $\hat{T}_\sigma$ are not rescaled, thus putting effectively more penalty on unimportant predictors, with little variance of $\hat{T}_\sigma$ across the samples $X_1,\ldots,X_n$ and less penalty on the important predictors with higher variance, see \citep{yuan05on} for details. 

Algorithmically, the problem can be solved with an efficient Lars algorithm  \citep{efron04least}, which can easily be adapted to include the positivity constraint. Alternatively, quadratic programming  can be used.

It might be surprising to see the $\ell_1$-norm constrained by 1 instead of a tuning parameter~$\lambda$. Yet this is indeed one of the interesting properties of Forest Garrote. The tree ensemble is in some sense selecting a good level of sparsity.  It seems maybe implausible that this would work in practice, but some intuitive reasons for its empirical success are given further below and  ample empirical evidence is provided in the section with numerical results.

 A drawback of the Garrote in the linear model setting is the reliance on the OLS estimator (or another suitable estimator), see also \cite{yuan05on}. The OLS estimator is for example not available if $p>n$. The tree ensemble estimates  are, in contrast, very reasonable estimators in a wide variety of settings, certainly including the high-dimensional setting $p\gg n$. 

The entire Forest Garrote algorithm works thus as follows
\begin{enumerate}
\item Fit Random Forest or another tree ensemble approach to the data. 
\item  Extract $\hat{T}_\sigma$ from the tree ensemble for all $\sigma\in\mathcal{G}$ by first extracting all rules $R_j$ and corresponding regression coefficients $\hat{\beta}_j$ and grouping them via (\ref{Tsigma}) for each interaction pattern.
\item Estimate $\hat{\gamma}$ as in (\ref{gammahat}) from the data, using for example the LARS algorithm \citep{efron04least}.
\item The fitted Forest Garrote function $\hat{T}^{gar}$ is given by (\ref{hatgar}).
\end{enumerate}
The whole algorithm is very simple and fast, as there is no tuning parameter to choose.
\subsection{Lack of tuning parameter}
In  most regularization problems, like the Lasso \citep{tibshirani96regression}, choosing the regularization parameter is very important and it is usually a priori not clear what a good choice will be, unless the noise level is known with good accuracy (and it usually it is not). The most obvious approach would be cross-validation. Cross-validation can be computationally expensive and is usually not guaranteed to lead to optimal sparsity of the solution, selecting many more variables or interaction patterns than necessary, as shown for the Lasso in \citet{meinshausen04consistent} and \citet{leng06nla}. 

Since the starting point is a very useful predictor, the original tree ensemble, there is a natural tuning parameter for the Forest Garrote estimator. 
As noted before, $\gamma=(1,1,1,\ldots,1,1)$ corresponds to the original tree ensemble solution. The original tree ensemble solution $\hat{T}$ is thus contained in the feasible region of the optimization problem (\ref{gammahat}) under a constraint on the $\ell_1$-norm of exactly 1.  The solution (\ref{hatgar}) will thus be at least as sparse as the original tree ensemble solution in the sense if sparsity is measured in the same $\ell_1$-norm sense as in (\ref{gammahat}).  Some variables might not be used at all even though they appear in the original tree ensemble.  

  On the other hand, the empirical squared error loss of $\hat{T}^{gar}$ is at least as low (and typically lower) than for the original tree ensemble solution as $\hat{\gamma}$ will reduce the empirical loss among all solutions in the feasible region of (\ref{gammahat}), which contains the original tree ensemble solution.
The latter point does clearly not guarantee better generalization on yet unseen data, but a constraint of 1 on the $\ell_1$-norm turns out to be an interesting starting point and is a very good default choice of the penalty parameter.

Sometimes one might still be interested in introducing a tuning parameter. One could replace the constraint of 1 on the $\ell_1$-norm of $\gamma$ in (\ref{gammahat}) by a constraint $\lambda$,
\begin{equation}\label{gammahatlambda} \hat{\gamma} \; =\; \mbox{argmin}_{\gamma}\; \sum_{i=1}^n (Y_i -  \sum_{\sigma\in\mathcal{G}} \gamma_\sigma  \hat{T}_\sigma(X_i))^2 \;\; \mbox{     such that     } \;\;|\mathcal{G}|^{-1} \sum_{\sigma\in\mathcal{G}} |\gamma_\sigma| \le \lambda \;\;\mbox{ and   } \min_{\sigma\in\mathcal{G}} \gamma_\sigma \ge 0.\end{equation}
 The range over which to search, with cross-validation, to over $\lambda$ can then typically be limited to $[0,2]$. 
Empirically, it turned out that the default choice $\lambda=1$ is very reasonable and actually achieves often better predictive and selection performance than the cross-validated solution, since the latter suffers from possibly high variance for finite sample sizes.

\begin{figure}
\begin{center}
\includegraphics[width=0.99\textwidth]{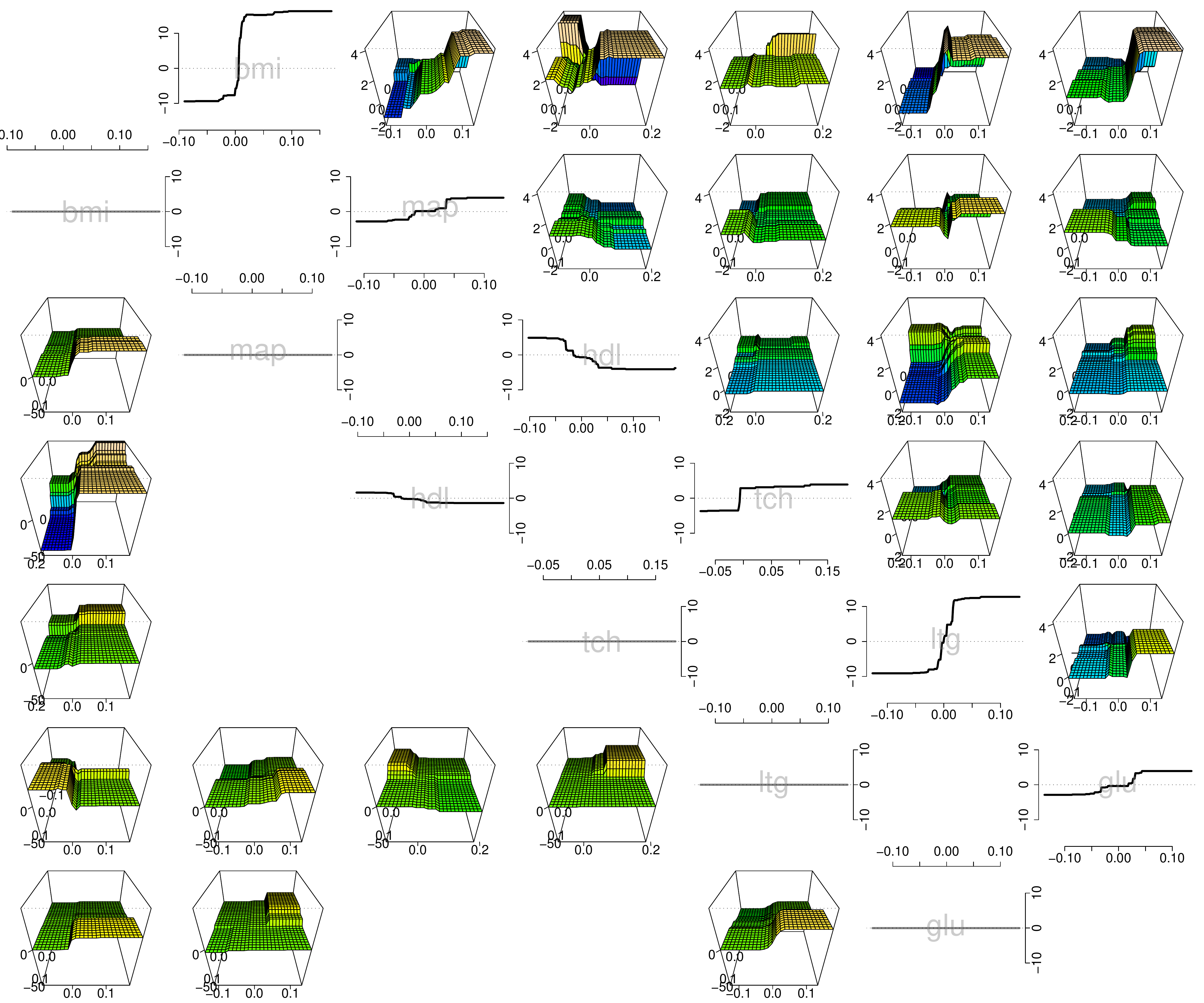}

\caption{\textit{ \label{fig:diabetes} \textit{ 
UPPER RIGHT DIAGONAL: `main effects' and `interactions' of second degree for the Random Forest fit on the Diabetes data between the main 6 variables (not showing all variables). LOWER LEFT DIAGONAL: corresponding functions for the Forest Garrote. Some main effects and interactions are set exactly to zero. Vanishing interactions are not plotted, leaving some entries blank. }
  }}
\end{center}
\end{figure}
\subsection{Example: diabetes data} The method is illustrated
on the diabetes data from \citet{efron04least} with $p=10$ predictor variables, age, sex, body mass index, average blood pressure 
and six blood serum measurements. These variables were obtained for each of $n = 442$ diabetes patients, along with the response of interest, a `quantitative measure of disease progression one year after baseline'. Applying a Random Forest fit to this dataset,  the main effects and second-order interactions effects, extracted as in (\ref{Tsigma}), are shown in the upper right diagonal of Figure~\ref{fig:diabetes}, for 6 out of the 10 variables (chosen at random to facilitate presentation). All of these variables have non-vanishing main effects (on the diagonal) and the interaction patterns can be quite complex, making them somewhat difficult to interpret. 

Now applying a Forest Garrote selection to the Random Forest fit, one obtains the main effects and interaction plots shown in the lower left diagonal of Figure~\ref{fig:diabetes}. Note that the x-axis in the interaction plots corresponds to the variable in the same column, while the y-axis refers to the variable in the same row. Interaction plots are thus rotated by 90 degrees between the upper right and the lower left diagonal. Some main effects and interactions are set to 0 by the Forest Garrote selection. Interaction effects that are not set to 0 are typically `simplified' considerably. The same effect was observed and explained in Figure~\ref{fig:38}. The interaction plot of Forest Garrote seems thus much more amenable for interpretation.

\section{Numerical Results}
\label{section:3}

\begin{figure}
\begin{center}

\includegraphics[width=0.99\textwidth]{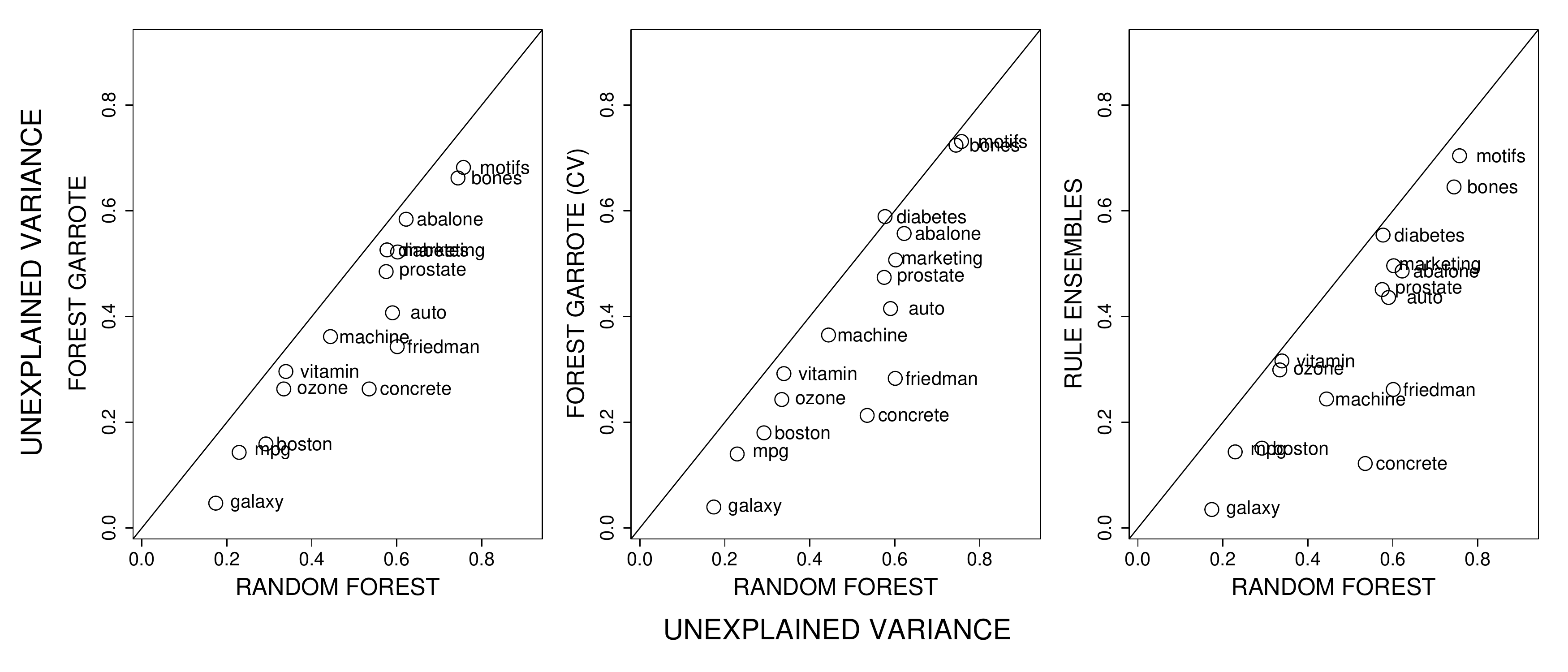}
\caption{ \textit{\label{fig:loss}  The unexplained variance on test data, as a fraction of the total variance. LEFT: Comparison of unexplained variance for Forest Garrote versus Random Forests. MIDDLE: Forest Garrote (CV) versus Random Forests. RIGHT: Rule Ensembles versus Random Forests. }}
\end{center}
\end{figure}

\begin{table}[h !]
\begin{center}
\begin{small}
\begin{tabular}{l l l |r r r} 

dataset & n & p & Forest Garrote & Forest Garrote (CV) & Rule Ensembles \\ \hline
motifs~~~&1294~~~~~&666~~~~~~&88.5~~~&425~~~&933~~~\\ 
ozone~~~~&165~~~~~~&9~~~~~~~~&~0.26~~~&~~2.7~~~&~16.5~~~\\ 
marketing&3438~~~~~&13~~~~~~~&~6.09~~~&~32.5~~~&696~~~\\ 
bones~~~~&242~~~~~~&3~~~~~~~~&~0.05~~~&~~1.1~~~&~26.2~~~\\ 
galaxy~~~&162~~~~~~&4~~~~~~~~&~0.07~~~&~~1.1~~~&~40.2~~~\\ 
boston~~~&253~~~~~~&13~~~~~~~&~0.79~~~&~~5.3~~~&~87.5~~~\\ 
prostate~&48~~~~~~~&8~~~~~~~~&~0.25~~~&~~3.3~~~&~~2.4~~~\\ 
vitamin~~&58~~~~~~~&4088~~~~~&~1.98~~~&~21.6~~~&~~39~~~\\ 
diabetes~&221~~~~~~&10~~~~~~~&~0.64~~~&~~4.8~~~&~31.2~~~\\ 
friedman~&150~~~~~~&4~~~~~~~~&~0.1~~~&~~1.3~~~&~16.6~~~\\ 
abalone~~&2088~~~~~&8~~~~~~~~&~0.59~~~&~~5.1~~~&808~~~\\ 
mpg~~~~~~&196~~~~~~&7~~~~~~~~&~0.18~~~&~~1.9~~~&~36.6~~~\\ 
auto~~~~~&80~~~~~~~&24~~~~~~~&~2.99~~~&~22~~~&~154~~~\\ 
machine~~&104~~~~~~&7~~~~~~~~&~0.29~~~&~~2.5~~~&~19.6~~~\\ 
concrete~&515~~~~~~&8~~~~~~~~&~0.58~~~&~~4.3~~~&271~~
\end{tabular}
\end{small}
\vspace{3mm}
\caption{\textit{ \label{table:TIMETABLE} The relative CPU time  spent on Forest Garrote, Forest Garrote (CV) and Rule Ensembles for the various datasets. Forest Garrote uses the least  computational resources since (i) it starts from a relative small set of dictionary elements (all $\hat{T}_\sigma$ for $\sigma\in \mathcal{G}$ as opposed to all rules), (ii) the solution has to be computed only for a single regularization parameter and there is hence (iii) also no need for expensive cross-validation. Note that the times above are only for the rule selection steps (\ref{ruleens}) and (\ref{gammahat}) respectively and the overall relative speed difference is typically smaller as a tree ensemble fit needs to be computed as an initial estimator in all settings. } }
\end{center}
\end{table}

To examine the predictive accuracy, variable selection properties and computational speed,  various standard datasets are used and augmented with two higher-dimensional datasets.
The first of these is a motif regression dataset (henceforth called `motif', $p=660$ and $n=2587$).
The goal of the data collection is to help find
transcription factor binding sites (motifs) in DNA 
sequences. The real-valued predictor
variables are 
abundance scores for $p$ candidate motifs (for each of the genes). Our
dataset is from a heat-shock experiment with yeast. For a general
description and motivation about motif regression see
\citet{conlon03motif}. 

The method is applied to a gene expression dataset (`vitamin') which is kindly
provided by DSM Nutritional Products (Switzerland). For $n=115$ samples, there
is a continuous response variable measuring the logarithm of riboflavin
(vitamin B2) production rate of Bacillus Subtilis, and there are $p=4088$
continuous covariates measuring the logarithm of gene expressions from
essentially the whole genome of Bacillus Subtilis. 
Certain mutations of genes are thought to lead to higher vitamin
concentrations and the challenge is to identify those relevant genes via regression, possibly using also interaction between genes.

In addition, the diabetes data from \citet{efron04least} (`diabetes', $p=10, n=442$), mentioned already above, are considered, the LA Ozone data (`ozone', $p=9, n=330$), and also the dataset about marketing (`marketing', $p=14, n=8993$), bone mineral density (`bone', $p=4, n=485$), radial velocity of galaxies (`galaxies', $p=4, n=323$) and prostate cancer analysis (`prostate', $p=9, n=97$); the latter all from \citet{hastie2001esl}. The chosen response variable is obvious in each dataset. See the very worthwhile book  \citet{hastie2001esl} for more details.
 To give comparison on more widely used datasets, Forest Garrote is applied to various dataset from the UCI machine learning repository \citep{UCI},  about predicting fuel efficiency (`auto-mpg', $p=8, n=398$),   compressive strength of concrete (`concrete', $p=9, n= 1030$),   median house prices in the Boston area (`housing' , $p=13, n=506$),   CPU performance (`machine', $p=10, n=209$) and finally the first of three artificial datasets in \citet{friedman91multivariate}.

The unexplained variance on test data for all these datasets with Forest Garrote is compared with that of Random Forests and Rule Ensembles in Figure~\ref{fig:loss}. The tuning parameters in Random Forests (namely over how many randomly selected variables to search for the best splitpoint) is optimized for each dataset, using the out-of-bag performance measure.
For comparison between the methods, the data are split into two parts of equal size, one half for training and the other for testing. Results are  compared with Forest Garrote (CV), where the tuning parameter $\lambda$ in (\ref{gammahatlambda}) is not chosen to be 1 as for the standard Forest Garrote estimator, but is instead chosen by cross-validation. There are two main conclusions from the Figure. First, all three method (Forest Garrote, Forest Garrote (CV) and Rule Ensembles) outperformed Random Forests in terms of predictive accuracy on almost all datasets. Second, the relative difference between these three methods is very small. Maybe surprisingly, using a cross-validated choice of $\lambda$ did not help much in improving predictive accuracy for the Forest Garrote estimator. On the contrary, it rather lead to worse predictive performance, presumably due to the inherent variability of the selected penalty parameter.

\begin{table}[h!]
\begin{center}
\begin{small}
\begin{tabular}{l l l | r r r r}
dataset & n & p & Forest Garrote &  Rule Ensembles & Random Forests\\ \hline
motifs~~~&1294~~~~~&666~~~~~~&44~~~~~~~&75~~~~~~~&233~~~~~~\\ 
ozone~~~~&165~~~~~~&9~~~~~~~~&~8~~~~~~~&~8~~~~~~~&~~9~~~~~~\\ 
marketing&3438~~~~~&13~~~~~~~&~9~~~~~~~&13~~~~~~~&~13~~~~~~\\ 
bones~~~~&242~~~~~~&3~~~~~~~~&~3~~~~~~~&~3~~~~~~~&~~3~~~~~~\\ 
galaxy~~~&162~~~~~~&4~~~~~~~~&~4~~~~~~~&~4~~~~~~~&~~4~~~~~~\\ 
boston~~~&253~~~~~~&13~~~~~~~&~9~~~~~~~&13~~~~~~~&~13~~~~~~\\ 
prostate~&48~~~~~~~&8~~~~~~~~&~8~~~~~~~&~8~~~~~~~&~~8~~~~~~\\ 
vitamin~~&58~~~~~~~&4088~~~~~&45~~~~~~~&67~~~~~~~&648~~~~~~\\ 
diabetes~&221~~~~~~&10~~~~~~~&~7~~~~~~~&~8~~~~~~~&~10~~~~~~\\ 
friedman~&150~~~~~~&4~~~~~~~~&~4~~~~~~~&~4~~~~~~~&~~4~~~~~~\\ 
abalone~~&2088~~~~~&8~~~~~~~~&~5~~~~~~~&~7~~~~~~~&~~8~~~~~~\\ 
mpg~~~~~~&196~~~~~~&7~~~~~~~~&~7~~~~~~~&~7~~~~~~~&~~7~~~~~~\\ 
auto~~~~~&80~~~~~~~&24~~~~~~~&16~~~~~~~&15~~~~~~~&~21~~~~~~\\ 
machine~~&104~~~~~~&7~~~~~~~~&~7~~~~~~~&~7~~~~~~~&~~7~~~~~~\\ 
concrete~&515~~~~~~&8~~~~~~~~&~8~~~~~~~&~8~~~~~~~&~~8~~~~~
\end{tabular}
\end{small}
\vspace{3mm}
\caption{\textit{\label{table:var} The number of variables selected in total for Forest Garrote, Rule Ensembles and Random Forests. Forest Garrote and Rule Ensembles prune the number of variables used considerably, especially for higher-dimensional data. } }
\end{center}
\end{table}

Forest Garrote (CV) has also obviously a computational disadvantage compared with the recommended Forest Garrote estimator, as shown in Table~\ref{table:TIMETABLE} which is comparing relative CPU times necessary to compute the relative estimators.
All three methods could be speeded up considerably by clever computational implementation. Any such improvement would most likely be applicable to any of these three compared methods as they have very similar optimization problems at their heart. Only relative performance measurements seem to be appropriate and only the time it takes to solve the respective optimization problems (\ref{ruleens}), (\ref{gammahat}) and (\ref{gammahatlambda}) is reported, including time necessary for cross-validation, if so required.  Rule Ensembles is faring by far the worst here, since the underlying optimization problem is very high-dimensional. The dimensionality $J$ in (\ref{ruleens}) is the total number of rules, which corresponds to the number of all nodes in the Random Forest fit. The total number $|\mathcal{G}|$ of interaction patterns in the optimization underlying (\ref{gammahat}) in the Forest Garrote fit is, on the other hand, very much smaller than the number $J$ of all rules, since many rules are typically combined in each interaction patterns. The lack of cross-validation for the Forest Garrote estimator clearly also speeds computation up by an additional factor between 5 and 10, depending on which form of cross-validation is employed.

Finally, the number of variables selected by either method is examined. A variable is said to be selected for this purpose if it appears in any node in the Forest or in any rule that is selected with a non-zero coefficient. In other words, selected variables will be needed to compute predictions, not selected variables can be discarded. The results are shown in Table~\ref{table:var}. Many variables are typically involved in a Random Forest Fit and both Rule Ensembles as well as Forest Garrote can cut down this number substantially. Especially for higher-dimensional data with large number $p$ of variables, the effect can be pronounced. Between Rule Ensembles and Forest Garrote, the differences are very minor with a slight tendency of Forest Garrote to produce sparser results.

\label{section:numerical}

\section{Conclusions}

Balancing interpretability and predictive power for regression problems is a difficult act. Linear models lend themselves more easily to interpretation but suffer often in terms of predictive power. Random Forests (RF), on the other hand, are known deliver very accurate prediction. Tools exist to extract marginal variable importance measures from RF. However, the interpretability of RF could be improved if the very large number of nodes in the hundreds of trees fitted for RF could be reduced.

Here, the Forest Garrote was proposed as such a pruning method for RF or tree ensembles in general. It collects all rules or nodes in the Forest that belong to the same functional group. Using a Garrote-style penalty, some of these functional groups are then shrunken to zero, while the signal of other functional groups is enhanced. This leads to a sparser model and rather interpretable interaction plots between variables. Predictive power is similar or better to the original RF fit for all examined datasets.

The unique feature of Forest Garrote is that it seems to work very well without the use of a tuning parameter, as shown on multiple well known and less well known datasets.  The lack of a tuning parameter makes the method very easy to implement and computationally efficient.

\end{document}